# Enhancing Health Data Interoperability with Large Language Models: A FHIR Study

Yikuan Li, MS[1,2,*], Hanyin Wang, BMed[1,*],
Halid Z. Yerebakan, PhD[2,†], Yoshihisa Shinagawa, PhD[2,†], Yuan Luo, PhD, FAMIA[1,†]
[1]Northwestern University Feinberg School of Medicine, Chicago, IL, U.S.A
[2]Siemens Medical Solutions, Malvern, PA, U.S.A

## Introduction

The integration and exchange of health data across diverse platforms and systems remain challenging due to the absence of standardized formats and a shared semantic understanding. This challenge becomes more significant when critical health information is embedded in unstructured data rather than well-organized structured formats. Standardizing unstructured health data, such as clinical notes, into FHIR resources can alleviate ambiguity across different health providers and, therefore, improve interoperability. However, it is by no means an easy task. Previous studies [1,2] have attempted to transform clinical notes into FHIR resources using a combination of natural language processing and machine learning tools through multi-step processes involving clinical named entity recognition, terminology coding, mathematical calculations, structural formatting, and human calibrations. However, these approaches require additional human effort to consolidate the results from multiple tools and have achieved only moderate performances, with F1 scores ranging from 0.7 to 0.9 in different elements. To this end, we intend to harness Large Language Models (LLMs) to directly generate FHIR-formatted resources from free-text input. The utilization of LLMs is expected to simplify the previously multi-step processes, enhance the efficiency and accuracy of automatic FHIR resource generation, and ultimately improve health data interoperability.

## Methods

**Data Annotation** To the best of our knowledge, there is no largely publicly available dataset in the FHIR standard that is generated from contextual data. Therefore, we have chosen to annotate a dataset containing both free-text input and structured output in FHIR formats. The free-text input was derived from the discharge summaries of the MIMIC-III datase. [3] Thanks to the 2018 n2c2 medication extraction challenge [4], which essentially involves named entity recognition tasks, elements in medication statements have been identified. Our annotations built upon these n2c2 annotations and standardized the free text into multiple clinical terminology coding systems, such as NDC, RxNorm, and SNOMED. We organized the contexts and codes into FHIR medicationStatement resources. The converted FHIR resources underwent validation by the official FHIR validator (https://validator.fhir.org/) to ensure compliance with FHIR standards, including structure, datatype, code sets, display names, and more. These validated results were considered the gold standard transformation results and could be used to test against the LLMs. No ethical concerns exist regarding data usage, as both the MIMIC and n2c2 datasets are publicly available to authorized users.

**Large Language Model** We used OpenAI's GPT-4 model as the LLM for FHIR format transformation. We used five separate prompts to instruct the LLM to transform input free text into medication (including medicationCode, strength, and form), route, schedule, dosage, and, reason, respectively. All prompts adhered to a template with the following strucuture: task instructions, expected output FHIR templates in .JSON format, 4-5 conversion examples, a comprehensive list of codes from which the model can make selections, and then the input text. As there was no fine-tuning or domain-specific adaptation in our experiments, we initially had the LLM generate a small subset (N=100). Then, we manually reviewed the discrepancies between the LLM-generated FHIR output and our human annotations. Common mistakes were identified and used to refine the prompts. It's important to note that we did not have the access to the whole lists of NDC, RxNorm, and SNOMED Medication codes for drug names, as well as SNOMED Finding codes for reasons. Additionally, even if we had such comprehensive lists, they would have exceeded the token limits for LLMs. Thus, we did not task LLMs with coding these entities; instead, we instructed them to identify the contexts mentioned in the input text. For other elements, e.g. drug routes and forms, numbering in the hundreds, we allowed LLMs to directly code them. When evaluating the LLM-generated output, our primary criterion was the exact match rate, which necessitates precise alignment with human annotations in all aspects, including codes, structures, and more. Additionally, we reported precision, recall, and F1 scores for specific element occurrences. We accessed the GPT-4 APIs through the Azure OpenAI service, aligning with responsible use guidelines for MIMIC data. The specific model we used was 'gpt-4-32k' in its '2023-05-15' version. Each text input was individually transformed into a MedicationStatement resource. To optimize efficiency, we made multiple asynchronous API calls.

---

*Co-first Authors; † Co-Senior Authors;
This study was conducted partially during Y. Li's summer internship at Siemens.

**Table 1.** Examples and Summary Statistics for Annotation Results and Performance of FHIR Conversion.

| Elements | Examples | N (%) | N, Uniq. Entries | Code System | N, Uniq. Codes | Exact Match | Prec. | Rec. | F1 |
|---|---|---|---|---|---|---|---|---|---|
| **medication** | | | | | | | | | |
| medicationCode | {'coding': [{'system': 'NDC', 'code': '51079088120', 'display': 'clonazepam 0.5 MG Oral Tablet'}, {'system': 'RxNORM', 'code': '197527','display': 'Clonazepam 500 microgram oral tablet'}, {'system': 'SNOMED', 'code': '322897008', 'display': 'Clonazepam 500 microgram oral tablet'}], 'text': 'clonazepam 0.5 mg Tablet'} | 3671 (100%) | 1762 | NDC RxNorm SNOMED | 625 520 210 | 0.968 | 1 | 1 | 1 |
| doseForm | {'text': 'Tablet','coding': [{'system': 'SNOMED','code': '385055001', 'display': 'Tablet'}]} | 1478 (40.3%) | 176 | SNOMED | 26 | 0.976 | 0.981 | 0.995 | 0.988 |
| totalVolume | {'value': 0.5, 'unit': 'milligram','system': 'http://unitsofmeasure.org','code': 'mg'} | 2383 (64.9%) | 188 | UoM | 16 | 0.980 | 0.968 | 0.978 | 0.973 |
| reason | [{'concept': {'text': 'headache', 'coding': [{'system': 'SNOMED', 'code': '25064002', 'display': 'Headache'}]}}] | 1106 (30.1%) | 619 | SNOMED | 354 | 0.902 | 0.986 | 0.943 | 0.964 |
| **dosage** | | | | | | | | | |
| route | {'text': 'PO', 'coding': [{'system': 'SNOMED', 'code': '26643006', 'display': 'Oral route'}]} | 2011 (54.8%) | 64 | SNOMED | 15 | 0.938 | 0.967 | 0.967 | 0.961 |
| timing.repeat | {'frequency': 1, 'period': 4.0, 'periodUnit': 'h', 'duration': 3.0, 'durationUnit': 'd'} | 2393 (65.2%) | 177 | hl7.org/fhir | 6 | 0.947 | 1 | 1 | 1 |
| timing.code | {'coding': [{'system': 'HL7','code': 'Q4H', 'display': 'Q4H'}]} | 2287 (62.3%) | 17 | hl7.org/fhir | 17 | 0.952 | 0.947 | 0.987 | 0.967 |
| doseQuantity /doseRange | {"doseQuantity": {"value": 5.0, "unit": "ML"}} | 1389 (37.8%) | 60 | | | 0.973 | 0.982 | 0.984 | 0.983 |

## Results and Discussions

The results of annotation and FHIR generation are presented in Table 1. In summary, we annotated 3,671 medication resources, covering over 625 distinct medications and associated with 354 reasons. The Large Language Model (LLM) achieved an impressive accuracy rate of over 90% and an F1 score exceeding 0.96 across all elements. In prior studies, F1 scores reached 0.750 in timing.repeat, 0.878 in timing.route, and 0.899 in timing dosage. [1] The LLM improved these F1 scores by at least 8%. It's worth noting that the previous studies used a smaller private dataset, did not employ the strictest evaluation metrics like exact match rate, skipped terminology coding, and required extensive training. On further investigation, we were also impressed by the high accuracy in terminology coding (which essentially involves a classification task with more than 100 classes), mathematical conversion (e.g., inferring a duration of 10 days when the input mentions '*TID, dispense 30 tablets*'), format conformity (with less than a 0.3% chance that the results cannot be interpreted in .JSON format), and cardinality (the LLM can handle both 1:N and 1:1 relationships).

The accuracy of the output is highly dependent on the instruction prompts used. Based on our extensive trials and errors, we have the following recommendations: i) provide diverse conversion examples that encompass a wide range of heterogeneous edge cases; ii) use strong language, such as "MUST", to ensure that the output adheres to the expected formats and rults; iii) continuously update and refine the prompts by reviewing results from a small subset, which can help identify common mistakes and enhance overall accuracy; iv) be cautious about out-of-vocabulary codings. LLMs may attempt to cater users by inventing codes that do not exist when they cannot find a close match.

## Conclusion

In this study, we provided the foundations of leveraging LLMs to enhance health data interoperability by transforming free-text input into the FHIR resources. Future studies will aim to build upon these successes by extending the generation to additional FHIR resources and comparing the performance of various LLM models.